\documentclass[sigconf]{acmart}

\usepackage{hyperref}
\usepackage{multirow}
\usepackage[ruled,vlined,linesnumbered]{algorithm2e}

\AtBeginDocument{%
  }

\copyrightyear{2022} 
\acmYear{2022} 
\setcopyright{acmcopyright}\acmConference[CIKM '22]{Proceedings of the 31st ACM International Conference on Information and Knowledge Management}{October 17--21, 2022}{Atlanta, GA, USA}
\acmBooktitle{Proceedings of the 31st ACM International Conference on Information and Knowledge Management (CIKM '22), October 17--21, 2022, Atlanta, GA, USA}
\acmPrice{15.00}
\acmDOI{10.1145/3511808.3557361}
\acmISBN{978-1-4503-9236-5/22/10}

\begin{document}

\title[Inductive Knowledge Graph Reasoning for Multi-batch Emerging Entities]{Inductive Knowledge Graph Reasoning for\\ Multi-batch Emerging Entities}

\settopmatter{authorsperrow=3}


\author{Yuanning Cui}
\author{Yuxin Wang} 
\author{Zequn Sun}
\affiliation{
    \department{State Key Laboratory for Novel Software Technology}
    \institution{Nanjing University \country{China}}
}
\email{{yncui, zqsun}.nju@gmail.com}
\email{yuxinwangcs@outlook.com}

\author{Wenqiang Liu}
\author{Yiqiao Jiang}
\author{Kexin Han}
\affiliation{
    \department{Interactive Entertainment Group}
    \institution{Tencent Inc \country{China}}
}
\email{masonqliu@tencent.com}
\email{gennyjiang@tencent.com}
\email{casseyhan@tencent.com}

\author{Wei Hu}
\authornote{Wei Hu is the corresponding author.}
\affiliation{
    \department{State Key Laboratory for Novel Software Technology}
    \department{National Institute of Healthcare\\ Data Science}
    \institution{Nanjing University \country{China}}
}
\email{whu@nju.edu.cn}

\renewcommand{\shortauthors}{Yuanning Cui et al.}

\begin{abstract}
Over the years, reasoning over knowledge graphs (KGs), which aims to infer new conclusions from known facts, has mostly focused on static KGs. 
The unceasing growth of knowledge in real life raises the necessity to enable the inductive reasoning ability on expanding KGs. 
Existing inductive work assumes that new entities all emerge once in a batch, which oversimplifies the real scenario that new entities continually appear. 
This study dives into a more realistic and challenging setting where new entities emerge in multiple batches. 
We propose a walk-based inductive reasoning model to tackle the new setting. 
Specifically, a graph convolutional network with adaptive relation aggregation is designed to encode and update entities using their neighboring relations. 
To capture the varying neighbor importance, we employ a query-aware feedback attention mechanism during the aggregation.
Furthermore, to alleviate the sparse link problem of new entities, we propose a link augmentation strategy to add trustworthy facts into KGs. 
We construct three new datasets for simulating this multi-batch emergence scenario. 
The experimental results show that our proposed model outperforms state-of-the-art embedding-based, walk-based and rule-based models on inductive KG reasoning.
\end{abstract}



\maketitle

\section{Introduction} 
\label{sect:introduction}

Knowledge graphs (KGs) are collections of massive real-world facts. 
They play a vital role in many downstream knowledge-driven applications, such as question answering and recommender systems \cite{Survey1,Survey2,Survey3}. 
Reasoning over KGs aims to discover new knowledge and conclusions from known facts.
It has become an essential technique to boost these applications.
In the real world, KGs like Wikidata \cite{Wikidata} and NELL \cite{NELL} are evolving, and unseen entities and facts are continually emerging \cite{OpenWorldKG}. From a practical point of view, this nature requires KG reasoning to be capable of learning to reason on continually-emerged entities and facts.

\begin{figure}[!t]
\centering
\includegraphics[width=.98\columnwidth]{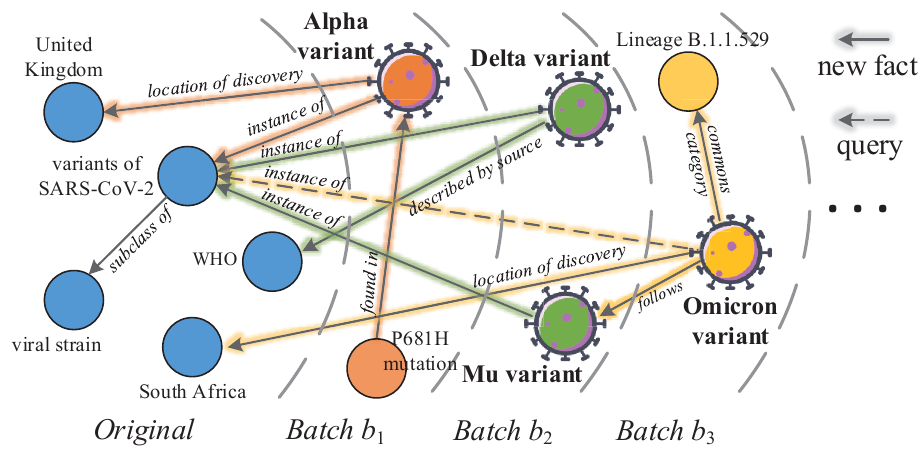}
\caption{An illustration of how new entities emerge in real life, excerpted from Wikidata \cite{Wikidata}. 
Solid glowing links are new facts as new entities emerge, while dashed glowing links are queries, e.g., (Omicron variant, \textit{instance of}, ?), to be inferred in the inductive KG reasoning task.}
\label{fig: MBE scenario}
\end{figure}

However, conventional KG reasoning models, e.g., \cite{TransE,TransH,DistMult,ConvE,InteractE,PairRE,KE-GCN,CompGCN}, are based on the closed-world assumption, where all entities must be seen during training. 
When new entities emerge, conventional models are unable to learn the embeddings for them unless re-training the whole KG from scratch. 
Although there have been some studies \cite{OOKB,Explainable,GraIL,INDIGO} to fill this deficiency, three critical defects of them in the settings put their actual reasoning ability into questions.
\emph{First}, existing work simulates only one single batch of emerging entities, which oversimplifies the real-world scenario where new entities are popping up continually. 
A real-world case extracted from Wikidata is illustrated in Fig.~\ref{fig: MBE scenario}, which depicts how new virus variants continually emerge.
\emph{Second}, newly emerging entities usually have sparse links \cite{GEN}. 
We observe in many existing benchmark datasets \cite{OOKB,LAN} that the average degree of emerging entities is even higher than that of entities in the original KG. 
See Table \ref{tab:degree} for example. 
\emph{Third}, some existing studies \cite{OOKB,LAN} do not consider emerging facts between emerging entities, e.g., the fact (P681H mutation, \textit{found in}, Alpha variant) in Fig.~\ref{fig: MBE scenario}, where both ``P681H mutation'' and ``Alpha variant'' newly appear in batch $b_1$.

\begin{table}[!t]
\centering
\caption{Comparison between the average degree of original entities in the KG and that of new entities over all emerging batch(es). The left four datasets are taken from two existing works, MEAN \cite{OOKB} and LAN \cite{LAN}, as examples, whereas the right three datasets are constructed in this work.}
\label{tab:degree}
\resizebox{\columnwidth}{!}{
    \begin{tabular}{lccccccc}
    \toprule \multirow{2}{1cm}{Avg. degree} & \multicolumn{2}{c}{MEAN} & \multicolumn{2}{c}{LAN} &  \multicolumn{3}{c}{Our MBE datasets}  \\
    \cmidrule(lr){2-3} \cmidrule(lr){4-5} \cmidrule(lr){6-8} & H-1000 & T-1000 & Sub-10 & Obj-10 & WN & FB & NELL   \\
    \midrule
    $\mathcal{E}_\text{Original}$ & 5.7 & 5.3 & 21.1 & 18.8  & 3.7 & 34.9 & 5.3 \\
    $\mathcal{E}_\text{New}$ & 9.5 & 11.8 & 63.1 & 72.2  & 2.0 & 14.4 & 1.2 \\
    \bottomrule
\end{tabular}}
\vspace{5pt}
\end{table}

Towards a more realistic modeling of emerging entities, we provide a re-definition of emerging entities and the inductive KG reasoning task in a \textbf{multi-batch emergence} (MBE) scenario. 
Unlike previous work, in this scenario new entities emerge in chronological multiple batches, and their degrees are restrained to make sure that they have relatively few links. 
Under this MBE scenario with fewer auxiliary facts for new entities, inductive KG reasoning becomes much more challenging. 

In this paper, we formulate the KG reasoning problem as reinforcement learning, and propose a new walk-based inductive model.
We aim to resolve three key challenges: 

\begin{enumerate}
\item \textbf{How to transfer knowledge to the emerging batches?}\\
Inductive learning aims at transferring knowledge learned from seen data to unseen data. 
To better exploit transferable knowledge, a walk-based agent is used to conduct the reasoning process in our model. 
It can learn to explicitly leverage the neighboring information, which is transferable to new entities. (Section~\ref{subsect:walk})

\item \textbf{How to encode new entities without re-training?}\\
We design ARGCN, a graph convolutional network (GCN) with adaptive relation aggregation, to not only encode new entities as they come, but also update the embeddings of existing entities that are affected by the new entities. (Section~\ref{subsect:ARGCN}) 
To attentively encode for specific queries, we devise a feedback attention mechanism which leverages logic rules extracted from reasoning trajectories to capture the varying neighbor importance of entities. (Section~\ref{subsect:feedback})

\item \textbf{How to resolve the link sparsity of new entities?}\\
Oftentimes, emerging entities only have a few links, which limits the walk-based reasoning. 
To alleviate this problem, we introduce a link augmentation strategy to add trustworthy facts into KGs for providing more clues. (Section~\ref{subsect:augment})
\end{enumerate}

Because there is no off-the-shelf dataset for the MBE scenario, we develop three new datasets in this work. 
The new entities in our datasets are divided into multiple batches and contain relatively sparse neighbors. 
Furthermore, our datasets contain not only emerging facts that link new entities with existing entities (a.k.a. unseen-seen facts), but also emerging facts that connect two new entities (a.k.a. unseen-unseen facts). (Section~\ref{sect:datasets})

We carry out extensive experiments on the new MBE datasets to compare with existing inductive models. 
The experimental results show that our model achieves the best reasoning performance on the inductive KG reasoning task. (Section~\ref{sect:experiment})

In summary, the key contributions of this paper are threefold:
\begin{itemize}
\item To the best of our knowledge, we are the first to explore inductive KG reasoning under the MBE scenario, which is more realistic and challenging.

\item We analyze the defects of the datasets in existing work, and construct new multi-batch datasets simulating real-world ever-growing KGs.

\item We propose a walk-based inductive KG reasoning model to cope with this new scenario. 
Our experimental results demonstrate the superiority of the proposed model against various kinds of state-of-the-art inductive models.
\end{itemize}

\section{Related Work} 
\label{sect:related_work}

In this section, we review existing embedding-based, walk-based and rule-based inductive KG reasoning models.

\subsection{Embedding-based Inductive KG Reasoning}

Conventional embedding-based KG reasoning models \cite{TransE,DistMult,ConvE,RSN} rest on static KGs. 
As aforementioned, these models are unable to handle emerging entities. 
As far as we know, MEAN \cite{OOKB} is the first inductive work that can learn the embeddings of emerging entities without re-training. 
It simply aggregated the transitioned neighboring information and took the mean results as the embeddings of entities. 
Based on MEAN, LAN \cite{LAN} incorporated the rule-based and neural attentions to measure the varying importance of different neighbors during aggregation. 
Both MEAN and LAN used a triple scoring function to rank each candidate fact.
A limitation of them is that they can only deal with unseen-seen facts due to their dependence on existing entities for aggregation.
GraIL \cite{GraIL} and TACT \cite{TACT} scored an extracted subgraph between two entities of each candidate fact.
They learned the entity-independent relational semantic patterns with graph neural networks (GNNs) to predict the missing relation for the entity pair.
INDIGO \cite{INDIGO} further exploited the structure information of KGs by fully encoding the annotated graphs which initialize entity features using their neighboring relation and node type information.

Other embedding-based work addressed specific settings or tasks of inductive KG reasoning. 
GEN \cite{GEN} and HRFN \cite{HRFN} explored the meta-learning setting where the edge degrees of entities are less than a small number.
PathCon \cite{PathCon} encoded each entity pair via aggregating its neighboring edges, which is similar to our ARGCN. 
Note that the query setting of PathCon is relation prediction in the form of $(h,?,t)$, rather than the harder entity prediction studied in this paper.
Since the settings and tasks of these studies cannot be directly adapted to the reasoning task under the MBE scenario, we do not discuss them in the rest of this paper.

\begin{figure*} 
\centering
\includegraphics[width=\textwidth]{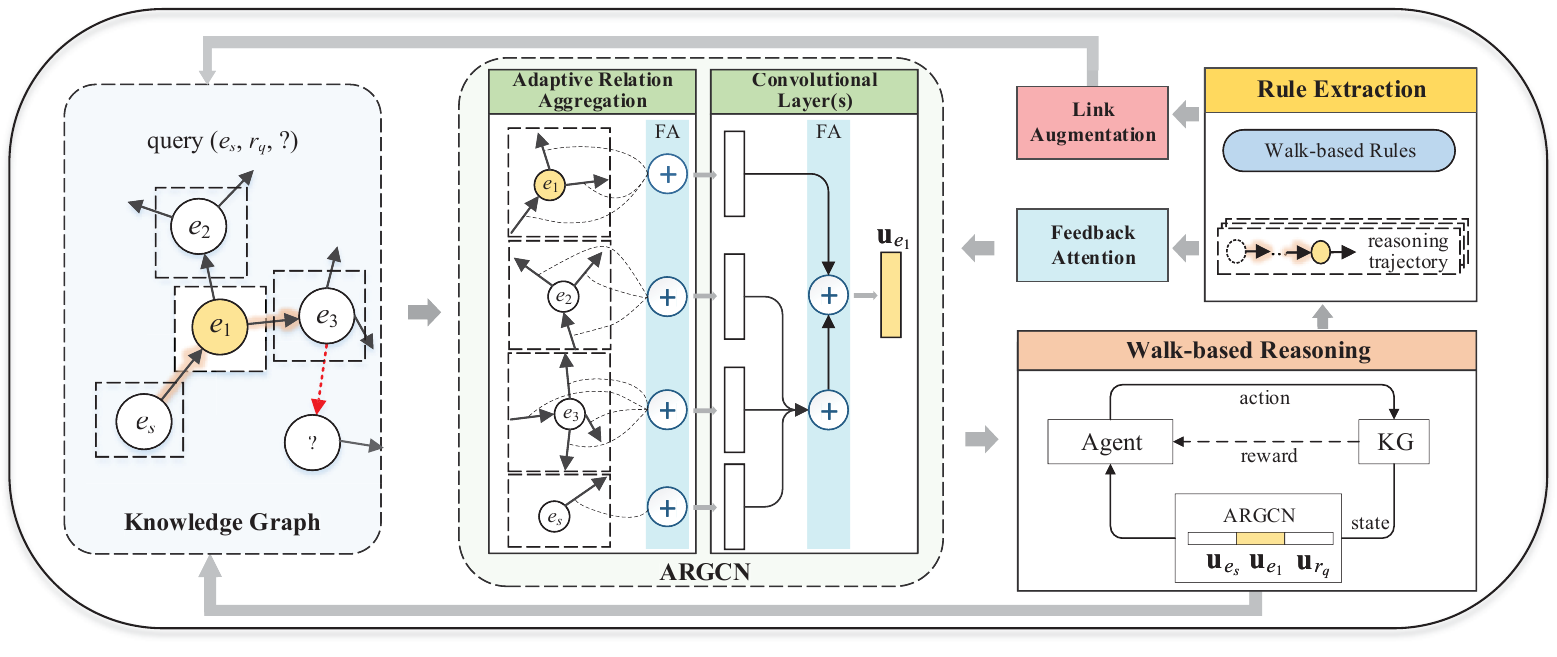}
\caption{Our model consists of four modules: walk-based agent, ARGCN, feedback attention and link augmentation.  
Given a query $(e_s, r_q, ?)$, we first use link augmentation to add trustworthy facts (e.g., the dotted red link from $e_3$ to $?$) into the KG.
Then, we use ARGCN with feedback attention to attentively encode entities and relations. 
Based on the embeddings, the walk-based agent conducts reasoning in the KG. 
Particularly, during training, our model extracts the walk-based rules from the reasoning trajectories to support feedback attention and link augmentation.}
\label{fig:overview}
\end{figure*}

\subsection{Walk-based Inductive KG Reasoning}

Instead of scoring candidate facts based on entity and relation embeddings, the KG reasoning problem can also be formulated in a reinforcement learning fashion, where a walk-based agent explores the reasoning paths to reach the target entities. 
DeepPath \cite{DeepPath} designed a KG-specific reinforcement learning environment and predicted the missing relations between entity pairs. 
MINERVA \cite{MINERVA} encoded the path history of an agent in a policy network and answered the missing tail entity of each query. 
Multi-Hop \cite{Multi-Hop} advanced MINERVA by introducing reward reshaping and action dropout to enrich the reward signals and improve generalization.
Other studies \cite{Meta-KGR,RLH,GaussianPath} proposed more complex reward functions, policy networks or reinforcement learning frameworks.
Particularly, GT \cite{Explainable} incorporated a graph Transformer for the walk-based reasoning and found that the walk-based models such as Multi-Hop can adapt well to the inductive setting, despite the emerging entities being randomly initialized.
RuleGuider \cite{RuleGuider} leveraged the pre-mined high-quality rules generated from AnyBURL \cite{AnyBURL} to provide more reward supervision. 
Both RuleGuider and our model leverage the rules to provide more knowledge for reasoning.
However, unlike RuleGuider, we do not need an extra rule mining model to obtain rules.

\subsection{Rule-based Inductive KG Reasoning}

Rule-based inductive models mine rules from KGs to help the prediction of missing facts.
Early rule mining models, e.g., AMIE \cite{AMIE} and AMIE+ \cite{AMIE+}, aimed at accelerating the mining process using parallelization and partitioning. 
They used specific confidence measures to discover the rules of high quality.
NeurLP \cite{NeurLP} and DRUM \cite{DRUM} learned the rules in an end-to-end differentiable manner, which computed a score for each fact and learned the rules by maximizing all fact scores. 
AnyBURL \cite{AnyBURL} extended the definition of rules to utilize more context information and devised a bottom-up method for efficient rule learning. 
The training process of walking-based models can also be regarded as rule mining, but their rules are implicit in the policy networks, rather than explicit in the form of Horn clauses.
In this paper, we extract explicit rules from the walk-based trajectories to assist reasoning over growing KGs. 

\section{Proposed Model} 
\label{sect:method}

In this section, we first introduce background concepts. 
Then, we present four key modules in our model, namely walk-based reasoning, ARGCN, feedback attention and link augmentation. 
The overview of our model is shown in Fig. \ref{fig:overview}.

\subsection{Preliminaries} 
\label{subsec:preliminaries}

\noindent\textbf{Expanding KG.}
In the MBE scenario, an expanding KG includes an existing KG called \textit{original} and $N$ \textit{batches} of emerging entities with concomitant facts:
\begin{itemize}
\item \textit{Original}. Let $\mathcal{K}_o = \{( e_h, r, e_t ) \,|\, e_h \in \mathcal{E}_o , r \in \mathcal{R}, e_t \in \mathcal{E}_o \}$ be the existing KG. 
$\mathcal{E}_o$, $\mathcal{R}$ are the entity and relation sets of $\mathcal{K}_o$, respectively. 
$(e_h, r, e_t)$ is called a fact, where $e_h, r$ and $e_t$ are the head entity, relation and tail entity, respectively.

\item \textit{Batches}. The $i$-th ($0\le i\le N$) emerging batch is denoted by $ \mathcal{K}_i = \{ (e_h, r, e_t) \,|\, r \in \mathcal{R}, e_h \in \mathcal{E}_i \lor e_t \in \mathcal{E}_i \}$, where $\mathcal{E}_i$ is the set of emerging entities appearing in this batch. 
$e_h$ and $e_t$ can both be emerging entities. 
Particularly, $\mathcal{K}_0=\mathcal{K}_o$. 
Note that we do not consider emerging relations in this paper.
\end{itemize}

\noindent\textbf{Inductive KG reasoning} aims to predict missing links for emerging entities. 
In this paper, each inductive KG reasoning query $q$ is a triple of $(e_s, r_q, ?)$, where $?$ denotes the target entity to be inferred. 
At least one of $e_s$ and $?$ in $q$ is an emerging entity.

\smallskip
\noindent\textbf{Walk-based reasoning} is a process that an agent starts from a head entity, continuously takes actions to move, and reaches a final entity. 
For an $r_q$-specific query $(e_s, r_q, ?)$, a reasoning trajectory is defined as a sequence of facts $ (e_s, r_1, e_1) \rightarrow (e_1, r_2, e_2) \rightarrow \cdots \rightarrow (e_{n-1}, r_n, e_\text{final}) $. 
We can extract a relation chain $r_1 \rightarrow r_2 \rightarrow \cdots \rightarrow r_n$ from the trajectory as a walk-based rule, denoted by $p$ for this $r_q$-specific query, i.e., $p: r_q\Leftarrow r_1\wedge r_2\wedge\ldots\wedge r_n$.
If the final entity $e_\text{final}$ is the target entity, this trajectory is called a \emph{positive} trajectory, otherwise it is called a \emph{negative} trajectory. 
We record the numbers of positive and negative trajectories that each $p$ resides in as $\text{pos}(p)$ and $\text{neg}(p)$, respectively. 
We denote the set of walk-based rules by $\mathcal{M}$ and the rule set for the $r_q$-specific query by $\mathcal{M}_{r_q}$.

\subsection{Walk-based Reasoning}
\label{subsect:walk}

We design a walk-based model to carry out the reasoning process. 

\smallskip
\noindent\textbf{Environment setting.}
We formulate the inductive KG reasoning as a Markov decision process (MDP) \cite{Multi-Hop} and train a reinforcement learning agent to perform the walk-based reasoning on the KG. 
A typical MDP is a quadruple $(\mathcal{S}, \mathcal{A}, \delta, \mathcal{B})$. 
In this paper, each element is defined as follows:
\begin{itemize}
\item\textit{States.}  
We define each state $s_t = \big((e_s, r_q), e_t\big)$ by combining the known part of query $(e_s, r_q, ?)$ and entity $e_t$ where the agent locates at step $t$. 
$\mathcal{S}$ is the set of all possible states.

\item\textit{Actions.}
In state $s_t$, the possible action space $\mathcal{A}_t  = \big\{ (r,e') \,|\,$ $(e_t, r, e') \in \mathcal{K}_o\big\} \cup \big\{ (r^-, e') \,|\, (e',r^-,e_t) \in \mathcal{K}_o \big\} \cup \big\{ (r_\text{sl}, e_t) \big\}$ is the union of neighboring edges of $e_t$ and a self-loop edge. 
$r^-$ denotes the reverse relation of $r$. 
The self-loop edge of an entity enables the agent to stay in the current state.

\item\textit{Transitions.}
The transition function $\delta : \mathcal{S} \times \mathcal{A} \rightarrow \mathcal{S}$ defines the transitions among states. 
Following an action $a_t \in \mathcal{A}_t$, the agent moves from $s_t$ to the next state $s_{t+1}$.

\item\textit{Rewards.}
After reaching the maximum number of moves, the agent stops on a final entity $e_l$ in state $s_l$. 
At that moment, a reward $\mathcal{B}$ is given to the agent, which is
\begin{align}
 \label{eq:reward}
    \mathcal{B} (s_l \,|\, e_s, r_q) =
    \begin{cases}
    1, & \mbox{if } (e_s, r_q, e_l) \in \mathcal{K}_o \\
    0, & \mbox{otherwise}
    \end{cases}.
\end{align}
\end{itemize}

\smallskip
\noindent\textbf{Policy network.}
To solve the MDP problem, the agent needs a policy to determine which action to take in each state.
We apply a neural network to parameterize the policy. 
In our policy network, the input consists of three parts: the embedding of $e_t$, the embedding of $r_q$ and the encoding of path history $(e_s, r_1, e_1, \ldots, r_t, e_t)$.
Here, we encode the path history with an LSTM as follows:
\begin{align}
    \mathbf{h}_0 &= \mathrm{LSTM}\big(\mathbf{0}, [\mathbf{u}_{r_0}; \mathbf{u}_{e_s}]\big), \\
    \mathbf{h}_t &= \mathrm{LSTM}\big(\mathbf{h}_{t-1}, [\mathbf{u}_{r_{t}};\mathbf{u}_{e_{t}}]\big),\quad t>0,
\end{align}
where $\mathbf{u}_{r_0} \in \mathbb{R}^d$ is the embedding of a special starting relation $r_0$, and $\mathbf{h}_t \in \mathbb{R}^{2d}$ is the encoding of the path history. 
Then, the policy network is defined as
\begin{align}
    \pi_\theta(a_{t}\,|\,s_t) = \sigma\Big(\mathbf{A}_t \times \mathbf{W}_{\mathrm{fc2}} \mathrm{ReLU} \big(\mathbf{W}_{\mathrm{fc1}} [\mathbf{u}_{e_t};\mathbf{u}_{r_q};\mathbf{h}_t]\big)\Big),
\end{align}
where $\mathbf{A}_t \in \mathbb{R}^{|\mathcal{A}_t| \times 2d}$ is the stacked embeddings of all actions. 
$\sigma()$ is the softmax operator. 
$\mathbf{W}_{\mathrm{fc1}} \in \mathbb{R}^{2d \times 4d}$ and $\mathbf{W}_{\mathrm{fc2}} \in \mathbb{R}^{2d \times 2d}$ are the learnable weight matrices of two fully-connected layers.

\smallskip
\noindent\textbf{Optimization.}
The objective of the policy network is to maximize the expected reward of all training queries:
\begin{align}
\label{eq:optimize}
    J(\theta) = \mathbb{E}_{(e_s,r_q,e_o) \in \mathcal{K}_o}\Big[\mathbb{E}_{a_1,\ldots,a_l \sim \pi_\theta}\big[\mathcal{B}(s_l \,|\, e_s,r_q)\big]\Big],
\end{align}
where $a_l, s_l$ are the action and state of the last step, respectively.

We use the REINFORCE optimization algorithm \cite{REINFORCE} to train our model, and the stochastic gradient is formulated as
\begin{align}
    \nabla_{\theta} J(\theta) \approx \nabla_{\theta} \sum_t \mathcal{B}(s_l\,|\,e_s,r) \log \pi_\theta (a_t \,|\, s_t).
\end{align}

It is worth noting that the embeddings of entities are also parameters to be learned. 
So, when new entities are added into the KG, the trained model cannot properly handle them since they do not have trained embeddings yet. 

\subsection{ARGCN} 
\label{subsect:ARGCN}

As just mentioned, the walk-based agent cannot encode emerging entities properly. 
To solve this problem, we propose a GCN with adaptive relation aggregation (ARGCN) as our encoder. 
Specifically, we first adopt an adaptive relation aggregation layer to aggregate the information of linked relations for each entity, and then use the stacked convolutional layers proposed in CompGCN \cite{CompGCN} to aggregate the information of multi-hop neighbors. 

\smallskip  
\noindent\textbf{Adaptive relation aggregation layer.}
Different from existing relational GNNs \cite{R-GCN,CompGCN} which randomly initialize the base embedding of each entity for training, we introduce an adaptive relation aggregation layer to learn the base embedding of each entity by solely aggregating its linked relations.

For an entity $e$ in $\mathcal{K}_o$, we generate its base embedding $\mathbf{u}_{e}^b \in \mathbb{R}^d$ as follows:
\begin{align} 
\label{eq:LS1}
    \mathbf{u}_e^b = \tanh\Big(
    \sum_{(r,e') \in \mathcal{N}_{\mathrm{in}}(e)} \mathbf{W}_{\mathrm{in}}\mathbf{z}_r + 
      \sum_{(r, e') \in \mathcal{N}_{\mathrm{out}}(e)} \mathbf{W}_{\mathrm{out}} \mathbf{z}_r
    \Big),
\end{align}
where $\mathcal{N}_{\mathrm{in}}(e)=\big\{(r,e')\,|\,(e',r,e)\in\mathcal{K}_o\big\}, \mathcal{N}_{\mathrm{out}}(e)=\big\{(r,e')\,|\, (e,r,$ $e')\in\mathcal{K}_o\big\}$. 
$\mathbf{W}_{\mathrm{in}}, \mathbf{W}_{\mathrm{out}}\in \mathbb{R}^{d\times d}$ are two learnable weight matrices. 
$\mathbf{z}_r \in \mathbb{R}^d$ is the learnable embedding of relation $r$ initialized with the Xavier normal initialization. 
$\tanh()$ is employed as the activation function.

\smallskip
\noindent\textbf{Stacked convolutional layers.}
We use a stacked multi-layered GCN to aggregate the multi-level neighboring information. 
The entity and relation embeddings are updated as follows:
\begin{align}
\label{eq:conv1}
\begin{split}
    \mathbf{u}^{l+1}_e &= \tanh\Big(\mathbf{W}^l_\mathrm{self} \mathbf{u}_e^l + \sum_{(r,e')\in \mathcal{N}_\mathrm{in}(e)} \mathbf{W}_\mathrm{in}^l \phi \big(\mathbf{u}_{e'}^l, \mathbf{u}^l_{r}\big) \\ 
      &\qquad\qquad\qquad\quad +\sum_{(r,e')\in \mathcal{N}_\mathrm{out}(e)} \mathbf{W}_\mathrm{out}^l \phi \big(\mathbf{u}_{e'}^l, \mathbf{u}^l_{r}\big)\Big),
\end{split}\\
    \mathbf{u}_r^{l+1} &= \mathbf{W}_\mathrm{rel}^{l}\mathbf{u}_r^l,\end{align}
where $\mathbf{u}^{0}_{e} = \mathbf{u}^{b}_{e}, \mathbf{u}^{0}_{e'} = \mathbf{u}^{b}_{e'}, \mathbf{u}^{0}_{r} = \mathbf{z}_{r}$.
$\mathbf{W}^{l}_\mathrm{in}, \mathbf{W}^{l}_\mathrm{out}, \mathbf{W}^{l}_\mathrm{self} \in \mathbb{R}^{d \times d}$ are three learnable weight matrices for entity embedding update, 
and $\mathbf{W}^{l}_\mathrm{rel} \in \mathbb{R}^{d \times d}$ is a learnable weight matrix for relation embedding update. 
The transition function $\phi: \mathbb{R}^{d} \times \mathbb{R}^{d} \to \mathbb{R}^{d}$ receives relation-entity pairs as input and aggregates the messages from neighboring links. 
We use the element-wise product for $\phi (\mathbf{u}_{e'}, \mathbf{u}_{r}) = \mathbf{u}_{e'} \circ \mathbf{u}_{r}$.

\subsection{Feedback Attention}
\label{subsect:feedback}

Transferring knowledge from seen data to unseen data is crucial for inductive reasoning. 
To transfer the KG reasoning patterns to emerging batches, we introduce a feedback attention mechanism which leverages the walk-based rules to capture the importance of neighboring relations for ARGCN. 

First, we calculate the confidence value of each walk-based rule $p$ as follows:
\begin{align}
\label{eq:conf}
    \text{conf}(p) = \frac{\text{pos}(p)}{\text{pos}(p)+\text{neg}(p)}.
\end{align}

Then, for $r_q$-specific queries, we calculate the correlation between $r_q$ and each other relation $r_j$ in the KG. 
We collect the walk-based rules containing $r_j$, and take the maximum confidence value of these rules as the correlation between $r_q$ and $r_j$:
\begin{align}
    \text{corr}(r_q, r_j) = \max\big\{\text{conf}(p_i)\,|\,p_i\in \mathcal{M}_{r_q} \wedge r_j\in p_i\big\}.
\end{align}
If $r_j$ is not in any walk-based rules, the correlation is set to 0.

At last, the weight of each $r_j$ given the query relation $r_q$ is
\begin{align}
    \alpha_{r_j\,|\,r_q} = \lambda_{r_q}\,\text{corr}(r_q, r_j) + (1-\lambda_{r_q}),
\end{align}
where $\lambda_{r_q} = \tanh\big(\frac{1}{\epsilon}\sum_{p_i\in \mathcal{M}_{r_q}}\text{pos}(p_i)\big) \in (0,1)$ and $\epsilon$ is a fixed hyperparameter.
Here we leverage $\lambda_{r_q}$ as a reliability measure to adjust the attention because the confidence of walk-based rules may not be reliable in the first few training iterations due to the small number of reasoning trajectories.

Put it all together, the re-formulated Eqs.~(\ref{eq:LS1}) and (\ref{eq:conv1}) after incorporating the feedback attentions are
\begin{align}
 \begin{split}
     \mathbf{u}_{e}^{b} &= \tanh\Big(\sum_{(r,e') \in \mathcal{N}_{\mathrm{in}}(e)}\alpha_{r\,|\,r_q}\mathbf{W}_{\mathrm{in}}\mathbf{z}_{r} \\
     &\qquad\qquad\qquad\quad +\sum_{(r,e') \in \mathcal{N}_{\mathrm{out}}(e)} \alpha_{r\,|\,r_q} \mathbf{W}_{\mathrm{out}} \mathbf{z}_r\Big),
 \end{split}\\
 \begin{split}
    \mathbf{u}^{l+1}_e &= \tanh\Big(\mathbf{W}_{\mathrm{self}} \mathbf{u}_e^l + \sum_{(r,e')\in \mathcal{N}_{\mathrm{in}}(e)} \alpha_{r\,|\,r_q} \mathbf{W}_{\mathrm{in}}^l \phi \big(\mathbf{u}_{e'}^l, \mathbf{u}^l_{r}\big) \\ 
    &\qquad\qquad\qquad\quad + \sum_{(r,e')\in \mathcal{N}_{\mathrm{out}}(e)} \alpha_{r\,|\,r_q} \mathbf{W}_{\mathrm{out}}^l \phi \big(\mathbf{u}_{e'}^l, \mathbf{u}^l_{r}\big)\Big).
 \end{split}
\end{align}

We take the output of the last layer as the final embeddings and use them in the walk-based reasoning process. 
With new entities and facts emerging, ARGCN adaptively aggregates relations in the new KG to generate and update the embeddings.

\subsection{Link Augmentation}
\label{subsect:augment}

KGs are widely acknowledged as incomplete \cite{Survey1,Survey2,Survey3}. 
Especially, emerging entities often suffer from the sparse link problem.
Many links of emerging entities have not been added to the KG, which hinders the reasoning on them. 
For example, as shown in Fig.~\ref{fig:overview}, due to the sparse links between the start entity $e_s$ and the target entity ?, the walk-based agent can hardly answer this query. 
However, if we can complement the missing link between $e_3$ and ?, this query would be likely to be answered.
Motivated by this, we propose a link augmentation strategy to add trustworthy facts into the KG for providing more reasoning clues.

Specifically, the link augmentation consists of two stages: generate candidate links and identify trustworthy ones. 
For candidate link generation, there are two methods: one is combining each entity in $\mathcal{E}$ and each relation in $\mathcal{R}$ one by one; the other is only combining the head entities that exist in the current facts $\mathcal{K}$ and their relations. 
The time complexity of the first method is $O(|\mathcal{E}|\times|\mathcal{R}|)$, which is very time-consuming. 
In practice, we find that this would also introduce a lot of noises, since there are no negative samples in the walk-based learning process. 
So, we apply the second method, which only has a time complexity of $O(|\mathcal{K}|)$.

To identify the trustworthy ones from candidate links, we leverage the walk-based rules as prior knowledge to select the trustworthy tail entities. 
Inspired by the traditional rule-based models \cite{AMIE,AMIE+}, we use two metrics, \emph{confidence} and \emph{support}, to measure the trustworthiness of rules, and use it to identify trustworthy rules:
\begin{align}
\label{eq:trustworthy_rules}
    \mathcal{P} = \big\{p\in \mathcal{M} \,|\, \text{conf}(p) \geq \gamma_1 \wedge \text{pos}(p) \geq \gamma_2 \big\}.
\end{align}
where $\mathcal{P}$ is the set of trustworthy rules, $\gamma_1$ is the confidence threshold, and $\gamma_2$ is the support threshold.

Then, we use the rules in $\mathcal{P}$ to select trustworthy facts. 
Specifically, for each candidate $(e_s, r_q)$, we use the $r_q$-specific rules to select tail entities to constitute the trustworthy facts. 
The generated fact set $\mathcal{L}$ can be formulated as
\begin{align}
\label{eq:trustworthy_links}
    \mathcal{L} &= \Big\{(e_s, r_q, e_t) \,|\, \exists \big\{(e_s, r_1, e_1), (e_1, r_2, e_2), \ldots , (e_n, r_n, e_t)\big\} \in \mathcal{K} \notag\\
    &\qquad\qquad \wedge (r_1 \rightarrow r_2 \rightarrow \cdots \rightarrow r_n) \in \mathcal{P}_{r_q} \wedge r_q \in \mathcal{R} \Big\},
\end{align}
where $\mathcal{P}_{r_q}$ is a set of $r_q$-specific rules and $\mathcal{P}_{r_q} \subseteq \mathcal{P}$.

Finally, we add $\mathcal{L}$ into the KG at the start of reasoning. 
Note that some trustworthy facts are duplicates of facts in the KG, but we still add them into the KG as the special rule-based facts. 
When an original fact is masked in the training stage, the agent may directly walk to the target entity through a duplicate rule-based fact.
With new entities and facts emerging, the link augmentation also generates trustworthy facts for new queries.

\begin{algorithm}[!tb]
\caption{Training process}
\label{alg:Train}
\KwIn{Training set $\mathcal{T}$ of the original KG $\mathcal{K}_o$.}
\KwOut{Model parameters $\theta$ and walk-based rule set $\mathcal{M}$.}
Initialize $\theta; \mathcal{M}\leftarrow\emptyset$, and augmentation fact set $\mathcal{L}\leftarrow\emptyset$\; 
\For{$epoch$ $\leftarrow 1$ \KwTo $max\_epochs$}{
    Update $\mathcal{L}$ and feedback attentions based on $\mathcal{M}$\; 
    Update the embeddings based on ARGCN with feedback attentions\;
    \ForEach{query $(e_s, r_q)$ in $\mathcal{T}$}{
        \For{$step \leftarrow 1$ \KwTo $max\_steps$}{
        The agent selects an action from $\mathcal{T} \cup \mathcal{L}$\;
        The agent walks to the next entity\;
        }
    Calculate the loss using Eq.~(\ref{eq:reward})\;
    Update $\theta$ using Eq.~(\ref{eq:optimize})\;
    Update $\mathcal{M}$ based on the generated reasoning trajectories\;
    }
}
\end{algorithm}


\begin{table*}
\caption{Statistics of the three constructed MBE datasets. 
$\mathcal{T}, \mathcal{V}, \mathcal{E}_{o}, \mathcal{R}$ denote the sets of training facts, validation facts, entities and relations in the original KG, respectively. 
$\mathcal{S}_i, \mathcal{Q}_i, \mathcal{E}_i$ ($1\le i\le 5$) are the sets of emerging facts, query facts and new entities of the $i$-th batch, respectively.}
\centering
\resizebox{\linewidth}{!}{
    \begin{tabular}{lccccccccccccccccccc}
    \toprule
    \multirow{2}{*}{Datasets} & \multicolumn{4}{c}{Original} & \multicolumn{3}{c}{Batch $b_1$} &    \multicolumn{3}{c}{Batch $b_2$} & \multicolumn{3}{c}{Batch $b_3$} & \multicolumn{3}{c}{Batch $b_4$} & \multicolumn{3}{c}{Batch $b_5$} \\ 
    \cmidrule(lr){2-5} \cmidrule(lr){6-8} \cmidrule(lr){9-11} \cmidrule(lr){12-14} \cmidrule(lr){15-17} \cmidrule(lr){18-20} & $|\mathcal{T}|$ & $|\mathcal{V}|$ & $|\mathcal{E}_{o}|$ & $|\mathcal{R}|$ & $|\mathcal{S}_1|$ & $|\mathcal{Q}_1|$ & $|\mathcal{E}_1|$ & $|\mathcal{S}_2|$ & $|\mathcal{Q}_2|$ & $|\mathcal{E}_2|$ & $|\mathcal{S}_3|$ & $|\mathcal{Q}_3|$ & $|\mathcal{E}_3|$ & $|\mathcal{S}_4|$ & $|\mathcal{Q}_4|$ & $|\mathcal{E}_4|$ & $|\mathcal{S}_5|$ & $|\mathcal{Q}_5|$ & $|\mathcal{E}_5|$ \\
    \midrule
    WN-MBE & \ \ 35,426 & \ \ 8,858 & 19,361 & \ \ 11 & \ \ 5,678 & 1,352 & \ \ 3,723 & \ \ 6,730 & 1,874 & \ \ 4,122 & \ \ 7,545 & 2,054 & \ \ 4,300 & \ \ 8,623 & \ \ 2,493 & \ \ 4,467 & \ \ 9,608 & \ \ 2,762 & \ \ 4,514 \\
    FB-MBE & 125,769 & 31,442 & \ \ 7,203 & 237 & 18,394 & 9,240 & \ \ 1,458 & 19,120 & 9,669 & \ \ 1,461 & 19,740 & 9,887 & \ \ 1,467 & 22,455 & 11,127 & \ \ 1,467 & 22,214 & 11,059 & \ \ 1,471 \\
    NELL-MBE & \ \ 88,814 & 22,203 & 33,348 & 200 & \ \ 4,496 & \ \ \,853 & \ \ 4,488 & \ \ 5,411 & 1,059 & \ \ 6,031 & \ \ 6,543 & 1,277 & \ \ 7,660 & \ \ 7,667 & \ \ 1,427 & \ \ 9,056 & \ \ 8,876 & \ \ 1,595 & 10,616 \\
    \bottomrule
\end{tabular}}
\label{tab:dataset_stat}
\end{table*} 

\subsection{Training and Answering Pipeline}
We integrate the above modules together and present a pipeline of training process in Algorithm~\ref{alg:Train}.
We are given a set of facts as the training data. 
After initialization, for each training epoch, we first update the augmentation fact set and feedback attentions based on the rule set $\mathcal{M}$. 
Then, we update entity embeddings using ARGCN with feedback attention in Line 4. 
In Lines 6--10, we perform the walk-based reasoning and calculate the loss to update the parameters $\theta$ of our model. 
In Line 11, we update the rule set $\mathcal{M}$ using the generated reasoning trajectory. 
Finally, the output is the learned model parameters and the set of walk-based rules.

With new entities and facts coming, we freeze parameters and perform query answering on the new KG, in the same way as Lines 6--8 in Algorithm~\ref{alg:Train}. 

\section{Datasets}
\label{sect:datasets}

To conduct a realistic evaluation for inductive KG reasoning, we construct three MBE datasets based on WN18RR \cite{ConvE}, FB15K-237 \cite{FB237} and NELL-995 \cite{DeepPath}, and name them as WN-MBE, FB-MBE and NELL-MBE, respectively. 
Here, we describe the steps to construct our MBE datasets. 
Given a static KG $\mathcal{G}$ (e.g., NELL), we generate a MBE dataset consisting of an original KG $\mathcal{K}_o$ and five emerging batches. 
$\mathcal{K}_o$ contains a training set $\mathcal{T}$ and a validation set $\mathcal{V}$. 
Each emerging batch $b_i$ contains an emerging fact set $\mathcal{S}_i$ and a query set $\mathcal{Q}_i$. 
The construction steps are as follows:
\begin{itemize}
\item \textit{Seeding}. We randomly sample $n$ entities in $\mathcal{G}$ as the seeds, and use them to constitute $\mathcal{E}_o$.
 
\item \textit{Growing}. We treat the seed entities in $\mathcal{E}_o$ as root nodes and conduct the breadth-first-search in $\mathcal{G}$. 
During the traversal, we add each visited entity into $\mathcal{E}_o$ with a probability $p$. 
This step is repeated until $|\mathcal{E}_o|\ge \frac{|\mathcal{E}_{\mathcal{G}}|}{2}$. 
Then, the entities not in $\mathcal{E}_o$ form another set $\mathcal{E}_{n}$.

\item \textit{Dividing}. For each fact in $\mathcal{G}$, we add it into $\mathcal{K}_o$ if both of its entities are in $\mathcal{E}_o$. 
We divide $\mathcal{K}_o$ into a training set $\mathcal{T}$ and a validation set $\mathcal{V}$ by a ratio $4:1$ on fact size. 
$\mathcal{E}_{n}$ is evenly divided into five batches, which simulates the MBE scenario. 
Then, for each batch, its corresponding emerging facts are chosen out from $\mathcal{G}$. 
Next, these facts are divided into an emerging fact set $\mathcal{S}_i$ and a query set $\mathcal{Q}_i$ using the minimal-spanning tree algorithm. 
The facts appearing in the trees are classified into $\mathcal{S}_i$ and the ones not in the trees constitute $\mathcal{Q}_i$. 
Using this algorithm avoids emerging entities to appear in the query sets only.

\item \textit{Cleaning}. Lastly, a few extra operations are done to ensure: 
(1) no new relations appear in the five batches; 
(2) for each batch, it should not contain any later emerging entity (i.e., chronological correctness); 
and (3) $\mathcal{E}_{\mathcal{V}}\subseteq\mathcal{E}_{\mathcal{T}}$, $\mathcal{E}_{\mathcal{O}_i} \subseteq \mathcal{E}_{\mathcal{T}}$ $\cup\,\mathcal{E}_{\mathcal{S}_{1}} \cup \ldots \cup \mathcal{E}_{\mathcal{S}_{i}}$. 
Note that these operations may cause some isolated new entities, which are discarded.
\end{itemize} 

In implementation, we set $n = 5,000, p = 0.5$; $n = 4,000, p = 0.5$ and $n = 3,000, p = 0.6$ for constructing WN-MBE, FB-MBE and NELL-MBE, respectively. The construction steps can easily repeat to discretionary amounts of emerging batches. 

Table~\ref{tab:dataset_stat} depicts the statistics of the three datasets. 
Our datasets satisfy the characteristic that new entities have sparse links (see Table~\ref{tab:degree} in Section~\ref{sect:introduction}). 

\section{Experiments and Results}
\label{sect:experiment}

In this section, we assess the proposed model and report our experimental results. 
The source code and constructed datasets are available at \url{https://github.com/nju-websoft/MBE}.

\begin{table*}
\caption{Hits@1 results of tail entity prediction under 1 vs. all. The best Hits@1 scores in each batch are marked in \textbf{bold}.}
\centering
{\small
    \begin{tabular}{lccccccccccccccc}
    \toprule
    \multirow{2}{*}{Models}  & \multicolumn{5}{c}{WN-MBE} & \multicolumn{5}{c}{FB-MBE} &   \multicolumn{5}{c}{NELL-MBE} \\
    \cmidrule(lr){2-6}\cmidrule(lr){7-11}\cmidrule(lr){12-16} & $b_1$ &  $b_2$ & $b_3$ & $b_4$ & $b_5$ & $b_1$ & $b_2$ & $b_3$ & $b_4$ & $b_5$ & $b_1$ & $b_2$ & $b_3$ & $b_4$ & $b_5$ \\
    \midrule
    MEAN        & \ \ 2.3  & \ \ 2.0 & \ \ 0.5 & \ \ 0.6 & \ \ 0.2 & 17.9 & 19.7 & 19.4 & 17.0 & 17.5 &  \ \ 8.2 & \ \ 4.5 & \ \ 4.2 & \ \ 2.6 & \ \ 4.9 \\
    LAN          & \ \ 2.7 & \ \ 2.9 & \ \ 2.6 & \ \ 4.3 & \ \ 1.5 & 21.2 & 21.3 & 18.8 & 16.5 & 16.9 & 14.1 & 11.2 & 10.3 & \ \ 8.3 & \ \ 7.0 \\
    Multi-Hop  & 80.6 & 79.7 & 78.8 & 79.0 & 79.9 & 31.1 & 30.3 & 29.9 & 28.2 & 28.2 & 59.6 & 60.2 & 60.8 & 64.6 & 63.9 \\
    GT           & 82.0 & 82.4 & 82.9 & 81.8 & 81.1 & 31.7 & 31.2 & 30.6 & 29.2 & 29.9 & \textbf{61.5} & 61.0 & 60.9 & 64.3 & 64.2 \\
    RuleGuider         & 79.2 & 80.2 & 77.8 & 75.9 & 75.9 & 24.6 & 25.1 & 23.9 & 22.3 & 23.0 & 57.1 & 52.9 & 56.9 & 58.7 & 59.8 \\
    NeurLP           & 80.1 & 76.5 & 72.5 & 73.8 & 73.7 & 23.1 & 23.1 & 23.2 & 22.5 & 23.3 & 47.4 & 43.4 & 43.1 & 46.8 & 46.1 \\
    DRUM           & 81.9 & 79.0 & 75.6 & 76.4 & 76.7 & 24.0 & 24.2 & 24.4 & 24.0 & 25.1 & 47.5 & 43.2 & 43.3 & 47.2 & 46.6 \\
    AnyBURL           & 83.6 & 82.9 & 79.2 & 78.3 & 77.4 & 29.1 & 27.5 & 16.5 & 26.0 & 26.5 & 56.4 & 57.4 & 61.8 & 62.8 & 64.4 \\
    \midrule
    Ours          & \textbf{84.5} & \textbf{84.9} & \textbf{85.1} & \textbf{84.4} & \textbf{83.9} & \textbf{32.6} & \textbf{32.6} & \textbf{31.8} & \textbf{30.9} & \textbf{32.1} & \textbf{61.5} & \textbf{63.0} & \textbf{67.0} & \textbf{67.2} & \textbf{69.9} \\
\bottomrule
\end{tabular}}
\label{tab:main_resultsH1}
\end{table*}   

\begin{table*}
\caption{MRR results of tail entity prediction under 1 vs. all. The best MRR scores in each batch are marked in \textbf{bold}.}
\centering
{\small
    \begin{tabular}{lccccccccccccccc}
    \toprule
    \multirow{2}{*}{Models}  & \multicolumn{5}{c}{WN-MBE} & \multicolumn{5}{c}{FB-MBE} & \multicolumn{5}{c}{NELL-MBE}  \\
    \cmidrule(lr){2-6}\cmidrule(lr){7-11}\cmidrule(lr){12-16} & $b_1$ &  $b_2$ & $b_3$ & $b_4$ & $b_5$ & $b_1$ & $b_2$ & $b_3$ & $b_4$ & $b_5$ & $b_1$ & $b_2$ & $b_3$ & $b_4$ & $b_5$ \\ 
    \midrule
MEAN      & 13.1 & 10.3 & \ \ 8.2 & \ \ 7.1 & \ \ 6.2  & 29.0 & 31.3 & 30.5 & 28.3 & 28.0 & 16.5 & 11.0 & \ 9.1  & \ \ 7.9  & \ \ 8.5 \\
LAN       & 17.6 & 17.3 & 17.7 & 17.6 & 16.7 & 31.9 & 33.0 & 30.3 & 27.4 & 26.7 & 22.2 & 20.4 & 18.1 & 16.7 & 14.5  \\
Multi-Hop    & 83.4 & 82.9 & 82.2 & 82.3 & 83.1 & 40.6 & 40.1 & 39.4 & 37.9 & 38.2 & 65.9 & 65.9 & 69.4 & 71.9 & 72.4 \\
GT          & 83.9 &  85.1 & 84.9 & 84.2 & 84.6 & 41.6 & 40.3 & 39.7 & 39.1 & 39.2  & 66.0 & 67.0 & 70.6 & 71.8 & 72.6 \\
RuleGuider      & 82.9 & 83.0 & 82.1 & 81.3 & 81.5 & 33.4 & 34.2 & 32.4 & 30.6 & 31.4  & 64.5 & 62.0 & 66.8 & 69.1 & 70.1 \\
NeurLP          & 82.1 & 79.0 & 75.4 & 76.7 & 76.7 & 30.7 & 30.6 & 30.7 & 30.4 & 31.0  & 52.4 & 47.3 & 47.4 & 51.4 & 50.3 \\
DRUM          & 80.1 & 76.6 & 72.9 & 73.3 & 74.0 & 31.6 & 31.4 & 31.7 & 31.5 & 32.5  & 53.1 & 47.7 & 47.7 & 51.6 & 50.6 \\
AnyBURL      & 85.7 & 85.8 & 83.5 & 83.2 & 82.8 & 37.6 & 36.1 & 23.7 & 34.7 & 35.0  & 63.0 & 62.7 & 67.0 & 68.8 & 70.9 \\
    \midrule
Ours           & \textbf{86.6} & \textbf{86.8} & \textbf{86.6} & \textbf{86.9} & \textbf{86.4} & \textbf{42.6} & \textbf{42.4} & \textbf{42.0} & \textbf{40.1} & \textbf{41.5} & \textbf{66.2} & \textbf{67.2} & \textbf{71.9} & \textbf{72.7} & \textbf{75.4} \\
\bottomrule
\end{tabular}}
\label{tab:main_resultsMRR}
\end{table*}   


\subsection{Experiment Settings} 
\label{subsec:experiment_settings}

\noindent\textbf{Competitors.} 
We compare our model with ten inductive KG reasoning models, including four embedding-based models: MEAN \cite{OOKB}, LAN \cite{LAN} GraIL \cite{GraIL} and INDIGO \cite{INDIGO}); three walk-based models: Multi-Hop \cite{Multi-Hop}, GT \cite{Explainable} and RuleGuider \cite{RuleGuider}; and three rule-based models: NeurLP \cite{NeurLP}, DRUM \cite{DRUM} and AnyBURL \cite{AnyBURL}. 

\smallskip
\noindent\textbf{Implementation details.} 
Since some comparative models are not originally designed for multi-batch emerging entities, we do some particular modifications to enable them. 
For the embedding-based models, after they generate embeddings for one batch, we combine them into $\mathcal{K}_o$ for the next batch. 
For the walk-based models, when testing queries in the $i$-th batch, we conduct reasoning on $\mathcal{K}_o \cup \mathcal{S}_1 \cup \ldots \cup \mathcal{S}_i$. 
Note that Multi-Hop is not inherently inductive. 
Following \cite{Explainable}, we use the Xavier normal initialization to assign a random vector to each emerging entity as its embedding. 
As for the rule-based models, they mine rules in $\mathcal{T}$.
We use the official code to run all competing models. 
To run a model on the constructed datasets, we train it on $\mathcal{T}$ and validate it on $\mathcal{V}$. During testing, we freeze all parameters and only perform reasoning on $\mathcal{K}_o \cup \mathcal{S}_1 \cup \ldots \cup \mathcal{S}_i$.
In detail, we train 1,000 epochs for MEAN and LAN, 3,000 epochs for INDIGO, which are the same as those in their papers. 
We train 100 epochs for Multi-Hop, GT and our model. 
The model with the best validation performance is used for inductive reasoning on the test sets. 
All experiments are performed on a server with two NVIDIA RTX 3090 GPUs, two Intel Xeon Gold 5122 CPUs and 384GB RAM.


\smallskip
\noindent\textbf{Parameter configuration.}
During training, we tune the hyperparameters of competing models and our model with grid search. 
For MEAN and LAN, the range of entity and relation embedding dimension is \{50, 100\}, the learning rate is \{0.0005, 0.001, 0.01\}, the training batch size is \{512, 1,024\}, the neighbor number is \{15, 25, 35\} for FB-MBE and \{3, 5, 10\} for WN-MBE and NELL-MBE. 
For INDIGO, the hidden vector dimension is set to 64, the range of learning rate is \{0.0005, 0.001\}, and the dropout rate is \{0.2, 0.5\}. 
For Multi-Hop, GT and RuleGuider, the dimension of relations and entities is set to 100, the range of learning rate is \{0.001, 0.005, 0.001\}, the beam size is set to 128, the number of LSTM (i.e., history encoder) layers is set to 3, the range of maximum walk steps is \{3, 4, 5\}, the action dropout rate is set to 0.1 for WN-MBE, 0.5 for FB-MBE, and 0.3 for NELL-MBE. 
For RuleGuider, the rule threshold is set to 0.15. 
For our model, the settings are the same as the above walk-based models except the range of GCN layer number is \{1, 2, 3\}, and $\epsilon$ is set to 1,000.

\smallskip
\noindent\textbf{Evaluation metrics.}
Following the convention \cite{Multi-Hop,Explainable,RuleGuider}, we conduct the experiments on tail entity prediction. 
Two different comparison settings are considered. 
In the first setting (named ``1 vs. all''), we compare with MEAN, LAN and all walk-based and rule-based competitors.
These competitors, together with our model, calculate a score $f(e_s, r_q, e_c)$ for each candidate entity $e_c$, where the total pool of candidates is all seen entities including emerging entities in previous batches. 
For the walk-based models, we set the scores of unreachable entities to $-\infty$. 
In the second setting (named ``1 vs. 100''), we compare with GraIL and INDIGO. 
These two competitors need subgraph building and scoring to rank candidate entities for each test query, which are very time- and space-consuming. 
Thus, rather than using all seen entities as the candidate pool, we follow the setting of GraIL and INDIGO \cite{GraIL, INDIGO} and randomly choose 100 negative entities for each gold target entity in ranking.
Moreover, because FB-MBE has denser links which cause the subgraph building and scoring process very slow (e.g., GraIL takes about 154 seconds to answer one query, while INDIGO takes about 22 seconds), we do not compare on FB-MBE but only perform the experiments on WN-MBE and NELL-MBE.

Again, following the convention, we pick two metrics of entity prediction: Hits@1 and mean reciprocal rank (MRR). 
We take the filtered mean ranks of gold target entities to calculate Hits@1 and MRR. 
Larger scores indicate better performance.

\begin{table}[t!]
\caption{Hits@1 results of tail entity prediction of GraIL, INDIGO and our model under 1 vs. 100. The best scores are marked in \textbf{bold}.
}
\centering
\resizebox{\columnwidth}{!}{
    \begin{tabular}{lcccccccccc}
    \toprule 
    \multirow{2}{*}{Models}  & \multicolumn{5}{c}{WN-MBE} & \multicolumn{5}{c}{NELL-MBE} \\
    \cmidrule(lr){2-6}\cmidrule(lr){7-11} & $b_1$ & $b_2$ & $b_3$ & $b_4$ & $b_5$ & $b_1$ & $b_2$ & $b_3$ & $b_4$ & $b_5$ \\ 
    \midrule
    GraIL  & \textbf{94.5} & 94.3 & \textbf{95.2} & \textbf{95.3} & \textbf{95.4} & 43.4 & 51.1 & 62.1 & 68.1 & 73.5 \\
    INDIGO  & 17.9 & 18.2 & 19.3 & 21.2 & 21.4 & 49.2 & 47.7 & 46.8 & 47.1 & 47.2 \\
    \midrule
    Ours  & 94.2 & \textbf{94.8} & 93.4 & 94.2 & 93.8 & \textbf{79.1} & \textbf{78.7} & \textbf{84.1} & \textbf{86.1} & \textbf{87.4} \\
\bottomrule
\end{tabular}}
\vspace{5pt}
\label{tab:1vs100_Hits_1}
\end{table}
\begin{table}[t!]
\caption{MRR results of tail entity prediction of GraIL, INDIGO and our model under 1 vs. 100. The best scores are marked in \textbf{bold}.
}
\centering
\resizebox{\columnwidth}{!}{
    \begin{tabular}{lcccccccccc}
    \toprule 
    \multirow{2}{*}{Models}  & \multicolumn{5}{c}{WN-MBE} & \multicolumn{5}{c}{NELL-MBE} \\
    \cmidrule(lr){2-6}\cmidrule(lr){7-11} & $b_1$ & $b_2$ & $b_3$ & $b_4$ & $b_5$ & $b_1$ & $b_2$ & $b_3$ & $b_4$ & $b_5$ \\ 
    \midrule
    GraIL  & \textbf{95.7} & \textbf{95.8} & \textbf{96.4} & \textbf{96.5} & \textbf{96.5} & 56.8 & 63.3 & 72.3 & 77.6 & 81.3 \\
    INDIGO & 30.7 & 31.1 & 31.5 & 33.7 & 34.2 & 62.6 & 60.5 & 59.5 & 59.9 & 60.0 \\
    \midrule
    Ours  & 95.2 & 95.3 & 93.8 & 94.5 & 94.2 & \textbf{79.8} & \textbf{79.5} & \textbf{84.7} & \textbf{86.6} & \textbf{87.6} \\
\bottomrule
\end{tabular}}
\vspace{5pt}
\label{tab:1vs100_MRR}
\end{table}

\subsection{Main Results}
\label{subsec:results}

\noindent\textbf{1 vs. all.}
The inductive KG reasoning results under this setting are presented in Tables \ref{tab:main_resultsH1} and \ref{tab:main_resultsMRR}.
We observe that (1) our model outperforms all competitors on all emerging batches of the three datasets. 
(2) The embedding-based models MEAN and LAN have the worst Hits@1 and MRR results among all three types of competitors. 
Also, their performance on the sparse datasets WN-MBE and NELL-MBE is much worse compared to their performance on FB-MBE, which indicates the dependency of embedding-based models on dense links to well learn neighbor aggregators.
(3) Among the walk-based models Multi-Hop, GT, RuleGuider and ours,
Multi-Hop performs worse than GT and our model because it randomly initializes the embeddings of emerging entities and infers based solely on the query relations and neighboring links. 
GT leverages Transformer to handle emerging entities.
It achieves the second-best performance, and is only inferior to ours.
Although RuleGuider introduces more rewards into reinforcement learning than Multi-Hop, it performs worse. 
We believe that the deterioration is due to the two agent design of RuleGuider increases its dependence on well-learned entity embeddings, but RuleGuider can only randomly initialize new entity embeddings.
(4) NeurLP, DRUM and AnyBURL are naturally adaptable to the inductive setting. 
But the rule-based models have poorer ability on exploring extensive potential paths and capturing complex structural patterns, which limits their inference performance especially on the dense dataset FB-MBE.

\smallskip
\noindent\textbf{1 vs. 100.}
The Hits@1 and MRR results under this setting are shown in Tables \ref{tab:1vs100_Hits_1} and \ref{tab:1vs100_MRR}.  
As mentioned in Section~\ref{subsec:experiment_settings}, GraIL uses a very time- and space-consuming testing process. 
It extracts and utilizes subgraphs between start entities and candidate tail entities, which capture more structural information to calculate scores. 
Even so, the results of GraIL are still worse than ours on NELL-MBE and only slightly better than ours on WN-MBE.
From a high level, INDIGO and ARGCN in our model are similar in the way of dynamically generating base embeddings for entities. 
However, the performance of INDIGO is inferior to ours, especially on WN-MBE with only 11 relations. 
The reason for this is that the entity encoding dimension of INDIGO is fixed with the relation number, therefore it may not be trained sufficiently.

\begin{table}
\caption{Hits@1 results of ablation study.}
\centering
\resizebox{\columnwidth}{!}{
    \begin{tabular}{lcccccccccc}
    \toprule
    \multirow{2}{*}{Variants}   & \multicolumn{5}{c}{WN-MBE} & \multicolumn{5}{c}{FB-MBE}  \\
    \cmidrule(lr){2-6}\cmidrule(lr){7-11} & $b_1$ & $b_2$ & $b_3$ & $b_4$ & $b_5$ & $b_1$ & $b_2$ & $b_3$ & $b_4$ & $b_5$ \\ 
    \midrule
    Base model  & 80.6 & 79.7 & 78.8 & 79.0 & 79.9 & 31.1 & 30.3 & 29.9 & 28.2 & 28.2 \\
    \ + ARGCN       & 81.1 & 80.8 & 80.5 & 81.4 & 81.8 & 32.2 & 31.2 & 31.6 & 29.9 & 31.2 \\
    \ + ARGCN \& FA  & 82.2  & 81.3 & 81.1 & 81.6  & 82.2 & 32.5 & 31.9 & 32.5 & 30.2 & 31.4 \\
    \ + LA   & 84.1 & 84.6 & 84.7 & 84.2 & 83.5 & 31.9 & 30.5 & 30.3 & 28.4 & 29.1 \\
    \ Intact model  & 84.5 & 84.9 & 85.1 & 84.4 & 83.9 & 32.6 & 32.6 & 31.8 & 30.9 & 32.1 \\
\bottomrule
\end{tabular}}
\vspace{5pt}
\label{tab:ablation_study_hit1}
\end{table}

\begin{table}
\caption{MRR results of ablation study.}
\centering
\resizebox{\columnwidth}{!}{
    \begin{tabular}{lcccccccccc}
    \toprule
    \multirow{2}{*}{Variants}   & \multicolumn{5}{c}{WN-MBE} & \multicolumn{5}{c}{FB-MBE}  \\
    \cmidrule(lr){2-6}\cmidrule(lr){7-11} & $b_1$ & $b_2$ & $b_3$ & $b_4$ & $b_5$ & $b_1$ & $b_2$ & $b_3$ & $b_4$ & $b_5$ \\ 
    \midrule
    Base model   & 83.4 & 82.9 & 82.2 & 82.3 & 83.1 & 40.6 & 40.1 & 39.4 & 37.9 & 38.2 \\
    \ + ARGCN        & 83.6 & 84.4 & 83.3 & 83.9 & 84.1 & 41.8 & 41.9 & 41.4 & 39.9 & 40.7 \\
    \ + ARGCN \& FA   & 84.0  & 84.4 & 83.7 & 84.1  & 84.8 & 41.9 & 42.1 & 41.9 & 40.1 & 41.0  \\
    \ + LA   & 85.5 & 86.3 & 86.0 & 86.7 & 85.8 & 41.6 & 40.6 & 40.0 & 38.6 & 39.1  \\
    \ Intact model  & 86.6 & 86.8 & 86.6 & 86.9 & 86.4 & 42.6 & 42.4 & 42.0 & 40.1 & 41.5 \\
\bottomrule
\end{tabular}}
\vspace{5pt}
\label{tab:ablation_study}
\end{table}

\smallskip
\noindent\textbf{Ablation study.} 
To validate the effect of different modules in our model, we conduct an ablation study. 
Due to the space limitation, we only show the results on WN-MBE and FB-MBE in Tables~\ref{tab:ablation_study_hit1} and \ref{tab:ablation_study}.
Similar conclusions can be observed on NELL-MBE.
Specifically, we design four variants of our intact model: base model, ``+ ARGCN'', ``+ ARGCN \& FA'' (i.e., ARGCN with feedback attention), and ``+ LA'' (link augmentation). 
The base model is a vanilla multi-hop inductive model without any additional modules proposed in our model, and it is identical to the competitor Multi-Hop.
The later three variants add specific module(s) into the base model.
We observe that the three modules have different impacts on the sparse and dense datasets.
(1) Compared with the base model, other three variants with additional modules all achieve better reasoning performance. 
(2) Compared to the base model, ``+ ARGCN'' variant stabilizes the performance as more batches come. 
This shows that ARGCN can effectively handle new entities.
(3) ``+ ARGCN \& FA'' outperforms ``+ ARGCN'', which implies that the feedback attention module helps the aggregation layer focus more on important relations that are relevant to specific queries. 
(4) ``+ LA'' variant performs better than the base model especially on WN-MBE, which indicates the significance to conduct link augmentation for sparse KGs.

\begin{table}
\caption{Hits@1 results of replacing the walk-based rules with the pre-mined rules by AnyBURL on WN-MBE.}
\centering
\resizebox{\columnwidth}{!}{
    \begin{tabular}{lccccc}
    \toprule
    Models & $b_1$ & $b_2$ & $b_3$ & $b_4$ & $b_5$ \\ 
    \midrule
    Our model w/ AnyBURL & 83.2 & 84.5 & 84.3 & 84.2 & 83.6 \\
    Our model w/ walk-based rules & 84.5 & 84.9 & 85.1 & 84.4 & 83.9 \\
\bottomrule
\end{tabular}}
\vspace{5pt}
\label{tab:premined_rules_hits1}
\end{table}

\begin{table}
\caption{MRR results of replacing the walk-based rules with the pre-mined rules by AnyBURL on WN-MBE.}
\centering
\resizebox{\columnwidth}{!}{
    \begin{tabular}{lccccc}
    \toprule
    Models & $b_1$ & $b_2$ & $b_3$ & $b_4$ & $b_5$ \\ 
    \midrule
    Our model w/ AnyBURL  & 84.8 & 86.3 & 86.0 & 87.0 & 86.3 \\
    Our model w/ walk-based rules & 86.4 & 87.2 & 87.4 & 87.3 & 86.9 \\
\bottomrule
\end{tabular}}
\vspace{5pt}
\label{tab:premined_rules}
\end{table}

\subsection{Further Analyses}

\begin{table*}
\caption{Hits@1 and MRR results over unseen-seen and unseen-unseen queries. The last row denotes the proportion of unseen-unseen queries in the test set of each batch.}
\centering
{\small
    \begin{tabular}{llccccccccccccccc}
    \toprule 
    \multirow{2}{*}{Metrics}& \multirow{2}{*}{Queries} & \multicolumn{5}{c}{WN-MBE} & \multicolumn{5}{c}{FB-MBE} & \multicolumn{5}{c}{NELL-MBE}  \\
    \cmidrule(lr){3-7}\cmidrule(lr){8-12} \cmidrule(lr){13-17} 
                          & & $b_1$ &  $b_2$ & $b_3$ & $b_4$ & $b_5$ & $b_1$ &  $b_2$ & $b_3$ & $b_4$ & $b_5$ & $b_1$ &  $b_2$ & $b_3$ & $b_4$ & $b_5$ \\ 
    \midrule
    \multirow{2}{*}{Hits@1}
    & Unseen-seen  & 84.3 & 85.1  & 86.2  & 85.8   & 85.0 & 32.3 & 32.2 & 31.7 & 30.9 & 31.9  &  57.7 & 57.6 & 59.7 & 61.0 & 71.9 \\
    & Unseen-unseen & 85.4 & 83.8 & 77.5 & 74.1 & 73.6 & 38.1 & 41.1 & 33.3 & 30.6 &  36.2 & 91.7 & 83.5  & 85.2  & 81.3  & 65.7  \\
    \midrule
    \multirow{2}{*}{MRR}       
    & Unseen-seen  & 86.4 & 87.0 & 87.5 & 88.3 & 87.4 & 42.4 & 42.1 & 42.0 & 40.1 & 41.4 & 62.7 & 61.5 & 64.5  & 65.7 & 75.2  \\
    & Unseen-unseen & 87.4 & 86.0 & 80.5 & 76.9 & 76.8 & 46.5 & 49.1 & 41.1 & 40.2  & 44.0   & 93.8 & 88.5 & 90.5 & 88.5  & 75.9  \\
    \midrule
    \multicolumn{2}{c}{\% of unseen-unseen}  & 16.0 & 16.6 & 13.1 & 12.2 & \ \ 5.7 & \ \ 4.6 & \ \ 4.2 & \ \ 4.4 & \ \ 4.3 & \ \ 9.5  & 11.3 & 21.0 & 28.5 & 30.6 & 32.4 \\
\bottomrule
\end{tabular}}
\label{tab:u-u_results}
\end{table*}



\noindent\textbf{Incorporating pre-mined rules.} 
In this analysis, we replace the walk-based rules with pre-mined rules.
Specifically, we first use AnyBURL \cite{AnyBURL} as the rule miner to discover rules. 
Then, we use the found rules to generate feedback attentions and augmentation facts. 
Note that the pre-mined rules are not updated during training.
Tables~\ref{tab:premined_rules_hits1} and \ref{tab:premined_rules} show the results on WN-MBE. 
We see that our model can still gain satisfactory results, despite that the results are slightly lower than those of using the walk-based rules. The same conclusions can also be observed on the other two datasets which are not shown due to the lack of space.

\smallskip
\noindent\textbf{Investigation on unseen-unseen queries.}
As described in Section~\ref{sect:related_work}, some work \cite{OOKB,LAN} ignores the emerging fact $(e_h, r, e_t)$ such that both $e_h$ and $e_t$ are new entities. 
This motivates us to investigate the performance of our model over the unseen-seen and unseen-unseen queries, respectively. 
We calculate the proportion of unseen-unseen queries in the whole query set of each batch, and analyze the tail entity prediction results. 
We report the results on WN-MBE, FB-MBE and NELL-MBE in Table~\ref{tab:u-u_results}.
We can see that the results of unseen-unseen queries are even better than those of unseen-seen ones on some batches of the three datasets. 
The results validate the versatility and robustness of our model to tackle both unseen-seen and unseen-unseen queries.


\begin{figure}
\centering
\includegraphics[width=\columnwidth]{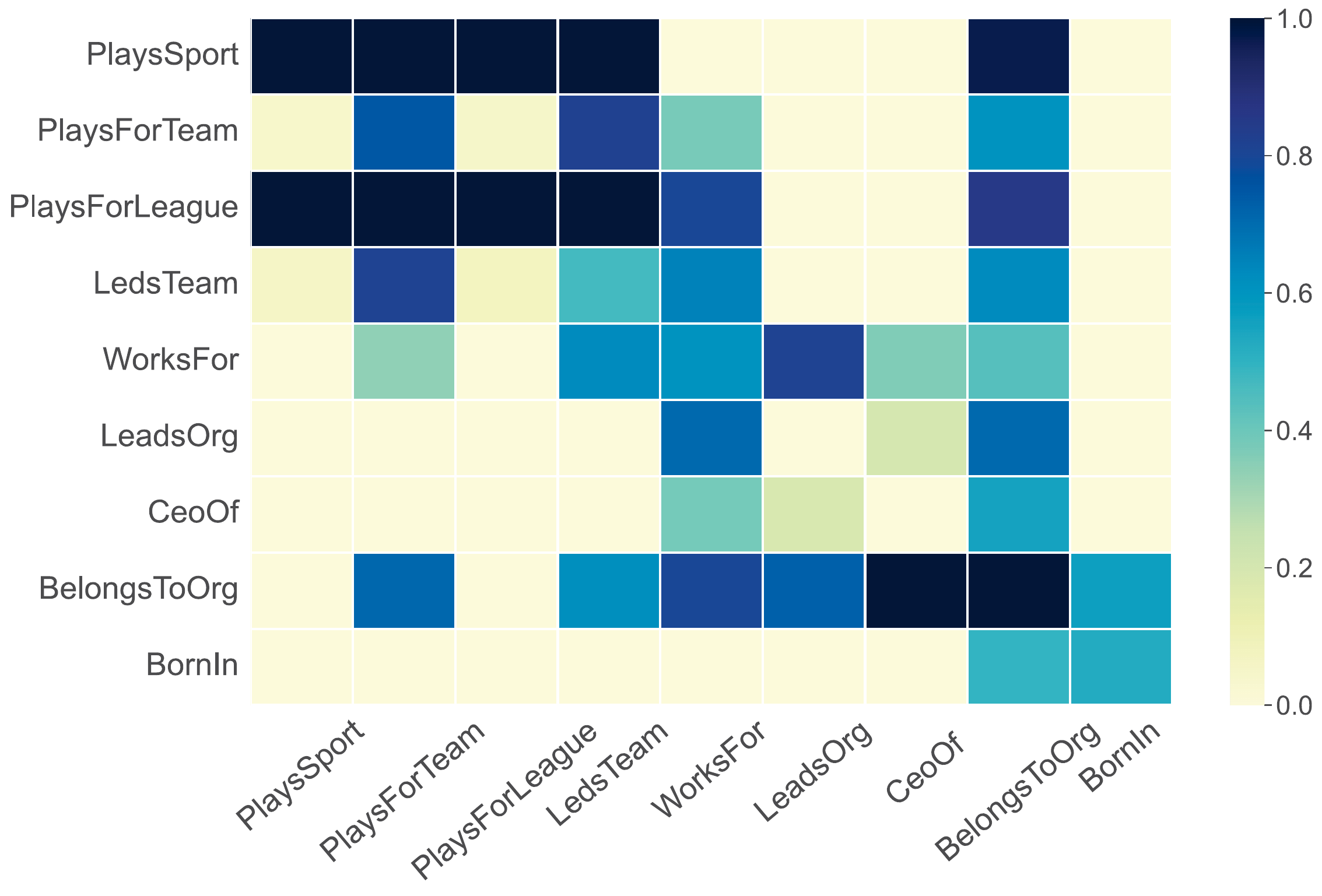}
\caption{Feedback attentions of nine person-related relations from NELL-MBE. Each horizontal axis tick label is a query relation, and the vertical axis tick labels are the relations to be aggregated.}
\label{fig: heat_map}
\end{figure}

\smallskip
\noindent\textbf{Heat map of feedback attentions.} 
In this experiment, we take insight into how feedback attentions capture the varying correlation of relations regarding specific queries.
In consideration of the readability, we choose several relations from NELL-MBE as examples for illustration. 
WN-MBE consists of conceptual-semantic and lexical relations of which the relation correlation is hard to measure, and the relations in FB-MBE are complex in structure and lack of readability. 
Specifically, we select nine person-related relations from NELL-MBE and show the feedback attentions between them in Fig.~\ref{fig: heat_map}.
The feedback attention mechanism encourages the agent to pay more attention to the query-related relations.
For example, when answering the query ($e_s$, \textit{play\_sport}, ?), the irrelevant relations such as \textit{born\_in} would be assigned a small attention.

\smallskip
\noindent\textbf{Case study.}
To demonstrate the reasoning ability of our model, we perform a case study on the trajectories that the agent walks.
We take three sport-related queries also from NELL-MBE as examples, and compare the reasoning trajectories of our model and the second-best model GT. 
For each case, we do beam search with the evaluated model and treat the probability of searching each trajectory as the score of this trajectory. 
Table~\ref{tab:case_study2} presents the top-ranked trajectories and their scores. 
Note that we remove the self-loop actions for clarity. 
It can be seen from the three cases that our model reaches all gold target entities, but GT fails in the first two cases, which shows the superiority of our model. 
In Case 2, although the trajectory is long, our model still succeeds because ARGCN with feedback attention focuses on the correlation between relations and can better learn complex relational path patterns. 
In Case 3, both our model and GT reach the target entity, but our model is more confident about this trajectory.
This is because the feedback attention is able to filter out noisy information for the query and make the action selection more stable.

\begin{table}
\caption{Case study. $r^-$ denotes the reverse relation of $r$. Target entities are \underline{underlined}.}
\centering
\resizebox{\columnwidth}{!}{
    \begin{tabular}{clc}
    \toprule
    Case 1. & (xabi, \emph{athlete\_home\_stadium}, \underline{stadium\_anfield}) & Scores \\
    \midrule
    \multirow{2}{*}{Ours} & $\to$ (\emph{athlete\_plays\_for\_team}, team\_liverpool) & \multirow{2}{*}{0.37} \\
                          & \quad\ $\to$ (\emph{team\_home\_stadium}, \underline{stadium\_anfield}) & \\
    \multirow{2}{*}{GT}   & $\to$ (\emph{athlete\_plays\_for\_team}, team\_liverpool) & \multirow{2}{*}{0.14} \\
                          & \quad\ $\to$ (\emph{athlete\_plays\_for\_team$^-$}, xabi) & \\
    
    \toprule
    Case 2. & (bowling\_green\_falcons, \emph{team\_plays\_in\_league}, \underline{ncaa}) & \\
    \midrule 
    \multirow{3}{*}{Ours} & $\to$ (\emph{agent\_collaborates\_with\_agent}, mexico\_ncaa) & \multirow{3}{*}{0.16} \\
                          & \quad\ $\to$ (\emph{agent\_controls}, unc\_wilmington\_seahawks) & \\ 
                          & \quad\ \quad\ $\to$ (\emph{team\_plays\_in\_league}, \underline{ncaa}) & \\
    GT              & $\to$ (\emph{agent\_collaborates\_with\_agent}, mexico\_ncaa) & 0.12 \\
    
    \toprule
    Case 3. & (brett\_carroll, \emph{athlete\_plays\_for\_team}, \underline{team\_bobcats}) & \\
    \midrule
    Ours            & $\to$ (\emph{athlete\_led\_sports\_team}, \underline{team\_bobcats}) & 0.17 \\
    GT              & $\to$ (\emph{athlete\_led\_sports\_team}, \underline{team\_bobcats}) & 0.04 \\
    \bottomrule
\end{tabular}}
\vspace{5pt}
\label{tab:case_study2}
\end{table} 

\section{Conclusion and Future Work} 
\label{sec:conclusion}

In this paper, we study a more challenging but realistic inductive KG reasoning scenario that new entities continually emerge in multiple batches and have sparse links. 
To address this new problem, we propose a novel walk-based inductive KG reasoning model. 
The experimental results on three newly constructed datasets show that our model outperforms the embedding-based, walk-based and rule-based competitors. 
In future work, we plan to study the reasoning task for expanding KG with emerging relations. 
We also want to investigate the effectiveness of our model in other applications such as multi-hop question answering. 

\smallskip\noindent\textbf{Acknowledgments.} 
This work was funded by National Natural Science Foundation of China (No. 62272219) and Collaborative Innovation Center of Novel Software Technology \& Industrialization.

\balance
\bibliographystyle{ACM-Reference-Format}
\bibliography{reference}

\end{document}